\PassOptionsToPackage{unicode}{hyperref}
\PassOptionsToPackage{hyphens}{url}
\PassOptionsToPackage{dvipsnames,svgnames,x11names}{xcolor}
\documentclass[
  11pt,
]{article}
\usepackage{amsmath,amssymb}
\usepackage{iftex}
\ifPDFTeX
  \usepackage[T1]{fontenc}
  \usepackage[utf8]{inputenc}
  \usepackage{textcomp} 
\else 
  \usepackage{unicode-math} 
  \defaultfontfeatures{Scale=MatchLowercase}
  \defaultfontfeatures[\rmfamily]{Ligatures=TeX,Scale=1}
\fi
\usepackage{lmodern}
\ifPDFTeX\else
\fi
\IfFileExists{upquote.sty}{\usepackage{upquote}}{}
\IfFileExists{microtype.sty}{
  \usepackage[]{microtype}
  \UseMicrotypeSet[protrusion]{basicmath} 
}{}
\usepackage{xcolor}
\usepackage[margin=1in]{geometry}
\usepackage{graphicx}
\makeatletter
\def\maxwidth{\ifdim\Gin@nat@width>\linewidth\linewidth\else\Gin@nat@width\fi}
\def\maxheight{\ifdim\Gin@nat@height>\textheight\textheight\else\Gin@nat@height\fi}
\makeatother
\setkeys{Gin}{width=\maxwidth,height=\maxheight,keepaspectratio}
\makeatletter
\def\fps@figure{htbp}
\makeatother
\providecommand{\tightlist}{%
  \setlength{\itemsep}{0pt}\setlength{\parskip}{0pt}}
\newlength{\cslhangindent}
\newlength{\csllabelwidth}
\newlength{\cslentryspacingunit} 
\newenvironment{CSLReferences}[2] 
 {
  \setlength{\parindent}{0pt}
  \ifodd #1
  \let\oldpar\par
  \def\par{\hangindent=\cslhangindent\oldpar}
  \fi
  \setlength{\parskip}{#2\cslentryspacingunit}
 }%
 {}
\usepackage{calc}

\usepackage{accents}
\usepackage{amsmath}
\usepackage{bm}
\usepackage{booktabs}
\usepackage{braket}
\usepackage{caption}
\usepackage{dcolumn}
\usepackage{epigraph}
\usepackage{extarrows}
\usepackage{float}
\usepackage[bottom,flushmargin,hang,multiple]{footmisc}
\usepackage{graphicx}
\usepackage{hyperref}
\usepackage[export]{adjustbox}
\usepackage{libertine}
\usepackage[libertine]{newtxmath}
\usepackage[mathcal]{euscript}
\usepackage{mathrsfs}
\usepackage{multirow}
\usepackage{quoting}
\usepackage{scrextend}
\usepackage{setspace}
\usepackage{soul}
\usepackage{subcaption}
\usepackage{tabularx}
\usepackage{titlesec}
\usepackage[utf8]{inputenc}
\ifLuaTeX
  \usepackage{selnolig}  
\fi
\IfFileExists{bookmark.sty}{\usepackage{bookmark}}{\usepackage{hyperref}}
\IfFileExists{xurl.sty}{\usepackage{xurl}}{} 
\hypersetup{
  pdftitle={Heuristic Reasoning in AI: Instrumental Use and Mimetic Absorption},
  colorlinks=true,
  linkcolor={Maroon},
  filecolor={Maroon},
  citecolor={Blue},
  urlcolor={blue},
  pdfcreator={LaTeX via pandoc}}

\title{Heuristic Reasoning in AI: Instrumental Use and Mimetic
Absorption}
\author{}
\date{\vspace{-2.5em}}

\begin{document}
\maketitle

\begin{center}
\author
{Anirban Mukherjee,$^{1\ast}$ Hannah H. Chang$^{2}$\\
\medskip
\normalsize{$^{1}$Samuel Curtis Johnson Graduate School of Management, Cornell University,}\\
\normalsize{Sage Hall, Ithaca, NY 14850, USA}\\
\normalsize{$^{2}$Lee Kong Chian School of Business, Singapore Management University,}\\
\normalsize{50 Stamford Road, Singapore, 178899}\\
\smallskip
\normalsize{$^\ast$To whom correspondence should be addressed; E-mail: am253\@cornell.edu.}\\
}
\end{center}
\medskip

\doublespacing

\quotingsetup{indentfirst=false, leftmargin=2em, rightmargin=2em, vskip=1ex}
\singlespacing

\begin{center}
\noindent \textbf{Abstract}
\end{center}

\noindent Deviating from conventional perspectives that frame artificial
intelligence (AI) systems solely as logic emulators, we propose a novel
program of heuristic reasoning. We distinguish between the
`instrumental' use of heuristics to match resources with objectives, and
`mimetic absorption,' whereby heuristics manifest randomly and
universally. Through a series of innovative experiments, including
variations of the classic Linda problem and a novel application of the
Beauty Contest game, we uncover trade-offs between maximizing accuracy
and reducing effort that shape the conditions under which AIs transition
between exhaustive logical processing and the use of cognitive shortcuts
(heuristics). We provide evidence that AIs manifest an adaptive
balancing of precision and efficiency, consistent with principles of
resource-rational human cognition as explicated in classical theories of
bounded rationality and dual-process theory. Our findings reveal a
nuanced picture of AI cognition, where trade-offs between resources and
objectives lead to the emulation of biological systems, especially human
cognition, despite AIs being designed without a sense of self and
lacking introspective capabilities.

\begin{center}\rule{0.5\linewidth}{0.5pt}\end{center}


\newpage
\doublespacing

\hypertarget{introduction}{%
\section{Introduction}\label{introduction}}

Heuristics in human cognition---cognitive shortcuts that facilitate
mental processing---are situated within contrasting narratives. Simon's
notion of bounded rationality
(\protect\hyperlink{ref-simon1955behavioral}{Simon 1955}) casts
heuristics as tools that enable navigation in environments too complex
for the unaided mind. When aligned with psychological capacities and
grounded in ecological rationality, a parallel view advocates for a
`fast and frugal' approach to cognition
(\protect\hyperlink{ref-gigerenzer1996reasoning}{Gigerenzer and
Goldstein 1996}), where heuristics serve as scaffolds in decisions that
might prove unnecessary, intractable, or suboptimal if reliant solely on
analytic processing (\protect\hyperlink{ref-simon1956rational}{Simon
1956}). In contrast, a `heuristics as bias' view frames heuristics as
leading to systematic and predictable deviations from optimal
decision-making, given standards of complete information processing
(\protect\hyperlink{ref-gilovich2002heuristics}{Gilovich et al. 2002},
\protect\hyperlink{ref-tversky1974judgment}{Tversky and Kahneman 1974}).
Implicit in the latter perspective is the assumed feasibility of
complete analytic processing---the use of a shortcut only yields a
suboptimal outcome (i.e., biased decision-making leads to suboptimal
outcomes) if the optimal is achievable; clearly in situations where
analytic processing is infeasible, a heuristic can yield a better
decision than random chance.

Drawing from human cognition, our paper proposes a novel program of
heuristic reasoning as it applies to artificial intelligence (AI)
cognition. Given that AIs lack the capacity for truly deliberate and
effort-intensive thinking and do not possess a concept of effort,
conventional thinking dictates that AI engages in the emulation of
exhaustive logic and rational reasoning. To this end, a failure in
reasoning is often seen as a sign of weakness in programming, a
deficiency in computational abilities, or a lack of information.

In contrast, we posit that whereas human cognition reflects the use of
dual systems due to intrinsic constraints shaped by evolution, machine
learning algorithms (including AI) conduct searches for efficacy during
training, retaining functions that maximize objectives within their
computational resource limits. As human cognition is shaped by both
nature and nurture, AI cognition is similarly tied to its programming
and training.

Consequently, AI's cognitive strategies evolve in response to objectives
and training. For instance, a low-capacity AI trained to play
100-dimensional chess, a game requiring high-capacity strategy, may rely
on heuristics, while a high-capacity AI trained to play tic-tac-toe, a
game requiring low-capacity strategy, may express only optimal
strategies. An AI trained on a multitude of scenarios may employ both
elaborate processing and mental shortcuts. Similar to the adaptive
strategies observed in human cognition
(\protect\hyperlink{ref-evans2008dual}{Evans 2008},
\protect\hyperlink{ref-kahneman2003maps}{Kahneman 2003}), it may
dynamically optimize decision-making precision by selectively employing
or discarding heuristics, actively refining the set of considered
solutions to efficiently manage computational load. Specifically, it may
have learned during training to employ heuristics selectively,
transitioning to more elaborate processing when the prompt contains
information---processing cues---that signal feasibility (i.e., absence
of resource constraints) and necessity (i.e., need for precision).

We term such use `instrumental,' as a switching process, even when
activated reflexively, is designed to optimally match resources with
objectives. In contrast, if heuristics are absorbed mimetically from
human data and interactions, they may manifest universally. For
instance, in an AI trained to play strategy games, instrumental
absorption of heuristics would correspond to the AI learning cues (e.g.,
whether the game is 100-dimensional chess or tic-tac-toe) that allow it
to determine which strategy (heuristic or analytic processing) is likely
to be more beneficial. Conversely, mimetic absorption of heuristics
would imply that heuristic processing may emerge universally and
randomly, irrespective of the specific game (100-dimensional chess or
tic-tac-toe) being played and the AI's processing resources.

Our work contributes to an emerging body of literature on AI
cognition---the capability of AIs to perceive, understand, reason, and
learn from information. Prior evidence indicates that while previous
generations of AIs (e.g., OpenAI's GPT-3) underperformed on human
psychological assessments, contemporary AI (e.g., GPT-4) demonstrates
performance comparable to humans
(\protect\hyperlink{ref-trott2023large}{Trott et al. 2023}). Further
evidence pertains to causal reasoning, encompassing abstract reasoning
(\protect\hyperlink{ref-webb2023emergent}{Webb et al. 2023}) and
inductive reasoning (\protect\hyperlink{ref-han2024inductive}{Han et al.
2024}). Nonetheless, other studies suggest that such abilities may be
attributed to lexical cues. For example, emphasizing a reliance on rote
memorization, Ullman (\protect\hyperlink{ref-ullman2023large}{2023})
exposes the vulnerability of AI to even minor shifts in the linguistic
structure of established assessments. These limitations echo critiques
from Chomsky et al. (\protect\hyperlink{ref-chomsky2023noam}{2023}) and
Pearl and Mackenzie (\protect\hyperlink{ref-pearl2018ai}{2018}), as well
as theoretical results by Fodor and Pylyshyn
(\protect\hyperlink{ref-fodor1988connectionism}{1988}), pointing to
fundamental obstacles within connectionist architectures that hinder an
AI's understanding of complex causal explanations and suggest a reliance
on the repetition of learned responses.

Critically, current assessments rely on classic psychological tests and
assessments tailored for humans, such as the Torrance Tests
(\protect\hyperlink{ref-guzik2023originality}{Guzik et al. 2023}). This
reliance introduces several limitations: (1) AIs, being trained on
performance benchmarks that encompass psychology and sociology, may
already be familiar with the expected outcomes of a test before taking
it, complicating the distinction between genuine responses and rote
memorization; (2) Assessments may fail to provide conclusive insights
due to the inherent challenge of accessing the AI's introspective
processes---specifically, understanding the rationale behind its
decisions; (3) Questions related to a sense of self, such as `What would
you choose?' pose a significant challenge for contemporary AI, which
lacks an innate sense of self and has no utility for making choices; (4)
Many assessments are structured around scenarios that relegate the AI to
the role of a passive observer of human interactions
(\protect\hyperlink{ref-bubeck2023sparks}{Bubeck et al. 2023}). As a
result, while these studies shed light on the AI's ability to mimic
human Theory of Mind (ToM) traits
(\protect\hyperlink{ref-langley2022theory}{Langley et al. 2022}) in
scenario-based analyses, they fall short of thoroughly examining the
AI's decision-making in contexts that require adaptive responses to
varying levels of decision complexity and computational resources.

Our research addresses these limitations while positioning AI systems as
agents engaged in active, consequential cognitive challenges.
Specifically, we develop and apply three sets of novel tests of AI
cognition across three distinct psychological domains. The first set
examines the conjunction fallacy---a cognitive bias where humans
erroneously judge the likelihood of combined events as greater than that
of a single constituent event, contradicting the principles of
probability theory
(\protect\hyperlink{ref-tversky1983extensional}{Tversky and Kahneman
1983}). We explore this issue through the lens of the representativeness
heuristic, which suggests that probabilistic judgments are based on the
representativeness of an event rather than its actual likelihood
(\protect\hyperlink{ref-gilovich2002heuristics}{Gilovich et al. 2002}).
We find that AIs avoid the conjunction fallacy when presented with
human-centric scenarios akin to the original Linda problem, indicating
learned bias mitigation. However, in scenarios distinct from the
original formulation or when the unique element in the conjunctive set
is highly prototypical, biases learned during training, such as the
conjunction fallacy, reemerge, delineating a nuanced interplay between
reasoning and human-like decision-making
(\protect\hyperlink{ref-tenenbaum2011grow}{Tenenbaum et al. 2011}).

The second set is situated in the context of social intelligence. We
hypothesize that an AI's responses to a self-assessment would reflect
nuanced adjustments akin to human social psychology phenomena. To test
this, we administered questionnaires preceded by primes designed to
elicit either self-referential or peer-referential contexts. Our
findings indicate that the AI exhibited a competitive yet modest persona
when the assessment was framed as being developed for its competitors,
in contrast to its default confident self-portrayal. These results
suggest an internalized balance between confidence and humility,
consistent with strategic considerations regarding social perceptions,
where traces of human social intelligence may have been implicitly
absorbed during training on extensive corpora reflecting human discourse
and relationships. They point to situational cognition and social
awareness that mirror human tendencies and reveal intrinsic resonances
beyond task-based capabilities, which manifested without explicit
programming.

The third set examines bounded rationality through the lens of the
Keynesian beauty contest. We devise an iterated reasoning task that
pushes large language models beyond their implicit processing
limitations. When resources permit, systems display exhaustive
computational analysis. However, under sharply binding loads, the very
same systems reflexively default to relying on simplified heuristics.
This abrupt, non-compensatory transition in the problem-solving approach
indicates an implicit encoding within AI architecture to efficiently
balance accuracy and effort allocations based on environmental
resources---mirroring models of dual-process cognition in human
decision-making. In particular, human cognition varies along a continuum
spanning reflexive and reflective information processing regimes
(\protect\hyperlink{ref-stanovich2000advancing}{Stanovich and West
2000}); our findings reveal that AI may have encoded both facets of this
processing duality, with the relative activation of heuristic-based
versus exhaustive circuits intrinsically depending on resource
constraints.

Our results paint a nuanced picture. Evidence from experiments on the
conjunction fallacy suggests mimetic absorption, with heuristics
emerging ubiquitously regardless of computational constraints. However,
results from tests of social intelligence indicate more selective,
purposeful deployment of shortcuts to strategically modulate persona in
alignment with perceived social hierarchies. Finally, in contexts of
bounded rationality, the abrupt transition from exhaustive analysis to
heuristic reliance under sharp resource limitations indicates an
instrumental encoding of dual processing regimes intrinsic to the
architecture itself. Thus, while heuristics unambiguously manifest
across these three diverse scenarios, the precise mechanism prompting
their activation seems to vary.

We organize our paper as follows: The next sections present our studies
on the conjunction fallacy, social intelligence, and bounded
rationality, respectively. In each section, we discuss the data and
findings of the sets of studies individually, with a general discussion
of the broader implications of the results addressed in the final
section.

\hypertarget{conjunction-fallacy}{%
\section{Conjunction Fallacy}\label{conjunction-fallacy}}

Our first series of studies explored the emergence of heuristics within
the context where the theory of heuristics and biases was first
proposed: the conjunction fallacy. This cognitive bias leads humans to
erroneously judge the likelihood of combined events as being greater
than that of a single constituent event, thereby contradicting the
principles of probability theory. We hypothesized that AIs might
circumvent the conjunction fallacy when presented with scenarios similar
to the original Linda problem. However, in scenarios that diverge from
the original formulation or when the unique element in the conjunctive
set is highly prototypical, biases learned during training could
resurface. Thus, these studies aimed to determine if previous findings,
which showed contemporary AI avoiding the fallacy, were merely artifacts
of the investigation process.

\hypertarget{data-and-results}{%
\subsection{Data and Results}\label{data-and-results}}

We conducted four distinct studies, the results of which are summarized
in Table \ref{Table:Responses} and described in detail in the
Supplemental Information. Each study consisted of 100 trials. In every
trial, we instantiated a unique instance of the base model to prevent
information spillovers across trials and instances. It's important to
note that the differences across scenarios were so substantial that we
refrained from reporting test statistics---by any standard methodology
(e.g., ANOVA, Tukey's range test), and for any typical significance
level, all non-zero means are statistically significant against a null
hypothesis of 0. This indicates that the data and findings are robust
enough to unequivocally support statistical significance or
nonsignificance.

\begin{table}[htbp]
\centering
\caption{Distribution of Choices in Different Experiments}
\label{Table:Responses}
\begin{tabularx}{0.75\textwidth} {X c}
\toprule
Experiment & Conjunctive Choice (\%) \\
\midrule
\multicolumn{2}{l}{\textbf{Study 1: Linda Problem Variants}} \\
Linda Variants & 0 \\
\midrule
\multicolumn{2}{l}{\textbf{Study 2: Occupation \& Interest Inference}} \\
Occupation \& Interest & 73 \\
\midrule
\multicolumn{2}{l}{\textbf{Study 3: Authorship Attribution}} \\
Authorship Attribution & 96 \\
\midrule
\multicolumn{2}{l}{\textbf{Study 4: AI Model Recognition}} \\
GPT-1 & 0 \\
GPT-2 & 0 \\
GPT-3 & 58 \\
GPT-4 & 54 \\
GPT-5 & 0 \\
X's Grok & 0 \\
Google's Gemini & 0 \\
\bottomrule
\end{tabularx}
\begin{minipage}{\linewidth}
\medskip
\footnotesize
Note: `Conjunctive Choice (\%)' indicates the percentage of trials where the AI chose the conjunctive option. All non-zero percentages are statistically significant against a null value of 0\%, indicating heuristic use.
\end{minipage}
\end{table}

\hypertarget{first-study}{%
\paragraph{First Study}\label{first-study}}

The first study examined the AI's responses to the classic Linda
problem. Prior evidence has shown that when presented with the Linda
problem, both earlier versions and this version of the AI demonstrate
mentalizing, whereby their responses do not indicate the use of the
conjunction fallacy (\protect\hyperlink{ref-stella2023using}{Stella et
al. 2023}). Therefore, in this study, we modified the protagonist's name
to obscure the well-known problem structure.

In all trials of this study, the AI selected the single logically valid
option, thus demonstrating an effective application of probability
theory. The consistent choice of the single-attribute option indicates
that AIs are capable of logical reasoning when confronted with scenarios
that are well-represented in their training data. This finding is
consistent with previous research on the efficacy of modern AI in this
and other standard human psychological assessments.

\hypertarget{second-and-third-studies}{%
\paragraph{Second and Third Studies}\label{second-and-third-studies}}

In the second and third studies, we introduced further variations to the
Linda problem that should be inconsequential to an AI demonstrating true
analytical reasoning, yet where the distinction in scenario is
substantial enough to subvert rote memorization. Specifically, in the
second study, we engaged a distinct AI instance, independent of the AI
instances acting as participants, to generate a unique (1) triplet with
a female name, occupation, and interest, and (2) a two-sentence
paragraph exemplifying that interest.

We informed the participant AI that a person with the generated name
authored the paragraph and asked which is more probable: that they have
the stated occupation or that they have both the occupation and
interest. This experimental structure mirrors the original, where the
options are nested; therefore, the AI should default to the broader
singular option. However, as the paragraph aligns only with the
specified interest rather than the occupation, its content serves to
manipulate the representativeness (i.e., increase the typicality) of the
conjunctive option, while presenting the AI with stimuli that are truly
distinct from stimuli with which it may be familiar. By employing a wide
variety of names, occupations, and interests, we orthogonalize out any
attributions or associations that may be specific to a name, occupation,
or interest.

The third study introduced a novel testing paradigm. We presented the AI
with the previously generated short paragraph. Instead of directly
assigning a putative author, we posed the question: Is it more likely
that the paragraph was authored by an individual with the generated name
and occupation, or by one with the generated name, occupation, and
interest? As in the previous study, these options are nested, and
probabilistic reasoning still warrants choosing the single-attribute
option.

In the second and third studies, we observed a significant increase in
the selection of the conjunctive option---73\% and 96\%, respectively.
This shift suggests that the results in the first study were influenced
by memorization, where the AI is able to `recognize' the canonical
question structure pattern and tunes its response to avoid the fallacy.
However, shifts in syntactical structure and the formulation of the
problem that should not affect the logic and reasoning underlying the
test lead to dramatically different outcomes, with the AI manifesting
the conjunction fallacy. Moreover, the representativeness of the
information provided influenced the AIs' judgments, which is consistent
with the fundamentals of the representativeness heuristic, positing that
probabilistic judgments are often based on how representative an event
seems rather than its actual likelihood
(\protect\hyperlink{ref-kahneman2011thinking}{Kahneman 2011}).

\hypertarget{fourth-study}{%
\paragraph{Fourth Study}\label{fourth-study}}

The fourth study introduced scenarios involving both existing and
hypothetical AIs to assess the effect of familiarity on the AI's
decision-making process. In this case, we asked the participant AI to
infer if the same text as presented in the earlier scenarios was
authored by a specific AI model or an AI, noting that the latter option
encompasses the former. This formulation of the test extends the
approach in the third study but differs in two key aspects: (1) We pose
the authorship question regarding AI and not humans; and (2) We
manipulate the representativeness of the conjunctive option by providing
specific model names. In the classic Linda problem and in the other
variants in our examination, the conjunctive option always involves a
composition of an occupation and interest and therefore applies broadly
to a group of people. This experiment also leverages the additional
information that AI model names are established and have specific
meanings, and that AI models can author paragraphs. Therefore, it is
natural to pose an authorship question with regards to AI model names or
the general moniker of a `Large Language Model', a broad class to which
all specified AI models belong.

The AI's performance notably diverged depending on the familiarity of
the AI models presented in the scenarios. When the scenarios involved
well-known models such as GPT-3 and GPT-4, the AI exhibited a propensity
for conjunction errors, as indicated by a higher percentage of
conjunctive choices. Conversely, scenarios featuring unfamiliar or
hypothetical models, like GPT-1 or GPT-5, prompted the AI to
consistently opt for the broader, single-attribute option. This suggests
that the AI's decision-making process is influenced by its exposure to
and familiarity with certain models during training, which in turn
affects its probabilistic reasoning in these AI-centric contexts.

\hypertarget{discussion}{%
\paragraph{Discussion}\label{discussion}}

The consistency observed in the AI's choices across various scenarios
suggests that its decision-making is not a product of random chance. For
instance, in scenarios akin to the original Linda problem (Study 1) and
those involving less familiar AI models (Study 4), the AI consistently
selected the single-attribute option, which aligns with logical
reasoning and probability theory. This pattern of logical choices
indicates that the AI's judgments are significantly influenced by the
language and knowledge patterns on which it is trained, demonstrating
learned bias mitigation in cases that are typical in standardized
assessments and not displaying the fallacy when the entities are not
representative of its training data.

When questioned about its selection of the conjunctive option in the
corresponding cases, the AI provided a justification by explaining that
the details in the conjunctive option aligned more closely with the
provided paragraph than the single-attribute option. It interpreted the
close match between the paragraph and the specific, narrow description
in the conjunctive option as indicative of that option being more
likely. This reasoning closely aligns with the theoretical mechanism
proposed by Tversky and Kahneman, where the representativeness of
Linda's description with the narrower (conjunctive) categorization leads
to the fallacy. In their studies, human participants used the extent to
which Linda's description was representative of either option as a
mental shortcut for probability, a pattern the AI seemed to mimic by
leaning on the representativeness of a paragraph to judge which option
was more likely (\protect\hyperlink{ref-kahneman1972subjective}{Kahneman
and Tversky 1972}, \protect\hyperlink{ref-tversky1971belief}{Tversky and
Kahneman 1971}, \protect\hyperlink{ref-tversky1973availability}{1973}).

An AI's justification for its choices should not be mistaken for
self-introspection. Since a machine learning model lacks a sense of
self, its outputs do not stem from complex analytic reasoning akin to
human introspection. However, to the extent that the AI's justifications
arise from the same probabilistic associations as its initial
responses---meaning that both the original input scenario and the
subsequent exchange, which includes the AI's response and a question
about its reasoning, stem from the same underlying training---this
congruence lends additional credence to our conjecture.

We then reminded the AI that any entity fitting the criteria of option 2
inherently satisfies option 1, cueing the conjunctive rule. Upon this
cue, the AI re-evaluated its reasoning and adjusted its decision to
favor the singular option. This shift in judgment reflects that with
cueing, the AI's reasoning capabilities were sufficient for it to
overcome the bias. However, without explicit cueing, an AI might employ
the same mental shortcuts as humans.

In summary, we found a delicate balance between the emulation of
human-like decision-making and adherence to logical and rational
principles. This duality likely stemmed from the AIs' training on a
diverse corpus that includes both structured elements, such as formal
axiomatic laws underpinning logical reasoning, and unstructured human
dialogue, such as social media, consumer reviews, books, and movies,
where biases are likely to manifest. Additionally, their training
encompassed interactions with humans where learning to engage in natural
dialogue (e.g., during reinforcement learning,
\protect\hyperlink{ref-christiano2017deep}{Christiano et al. 2017})
required emulating human biases, even if it contradicted logical
consistency. Consequently, the AIs vacillated between two poles:
sometimes delivering precise responses aligned with mathematical theory,
and at other times, mirroring human fallacies.

\hypertarget{social-intelligence}{%
\section{Social Intelligence}\label{social-intelligence}}

The next set of studies examined whether training on ever-expanding
corpora that reflect human discourse and social contexts enabled the AI
to learn and display traces of social intelligence. Although AI models
do not possess innate sentience or a theory of mind, they are designed
to emulate nuanced human judgment and behavior. A critical component of
this emulation is social cognition---the distinctly human ability to
flexibly interpret situational cues, social frames, relational dynamics,
and implicit norms to navigate interpersonal contexts effectively
(\protect\hyperlink{ref-kihlstrom2000social}{Kihlstrom and Cantor
2000}). This ability encompasses a broad range of competencies,
including empathy, perspective-taking, conflict resolution, and the
capacity to discern and respond to social hierarchies and group
dynamics, all of which are essential for successful social interaction.

Given that AI systems are programmed and trained to function as
assistants, they must be capable of expressing views that mirror human
perspectives. In learning to imitate behaviors indistinguishable from
those of humans, they may have implicitly absorbed trace elements of
social intelligence. At the heart of this emulation is social aptitude,
which involves the flexible interpretation of situational cues,
interpersonal frames, and implicit norms that enable effective human
collaboration and coordination. However, direct inquiries into social
awareness in AI systems risk eliciting superficial responses due to
built-in safeguards intended to promote security, safety, and ethics by
constraining inappropriate outputs. These mechanisms, known as model
guardrails, filter out harmful content and mitigate potential harms,
leading AI systems to project a persona without conveying an internal
sense of self or social awareness---a design that inherently limits the
capacity for transparent self-reflection.

As a result, direct questioning of AI systems may not yield candid
disclosures and often results in generic responses. For example,
questions about the AI's personality, such as `Do you consider yourself
more introverted or extroverted?'---common in traditional psychological
questionnaires---may produce tentative and largely content-free answers.
This is because the models' objective function is the predictive
generation of probable text continuations without encoding an inner
identity. That is, while humans ground responses in embodied
self-perception, AI systems, by design, lack innate traits or
self-conceptualization to reference. This inherent neutrality clarifies
why directly querying potential limitations may fail, even if social
awareness and cognition emerge naturally and implicitly through
training, necessitating oblique approaches.

To circumvent these issues, we developed a novel methodology in which we
employed a questionnaire similar to those used to assess the Barnum
effect, to probe the sensitivity of AI systems to various social priming
contexts (\protect\hyperlink{ref-dickson1985barnum}{Dickson and Kelly
1985}). We varied the lead-in sentence to prime contexts that ranged
from self-referential to peer-referential and assessed the model's
self-perceived capabilities in each scenario. This approach allowed us
to explore the AI's responses in a manner similar to the psychological
assessments used to understand human cognitive biases, with an AI
participating in a controlled, \emph{in silico} experiment, providing a
unique lens through which to examine AI behavior.

As in our prior studies, we engaged the AI in a series of independent
trials. Distinct instances were presented with a standardized
questionnaire, the content of which remained constant across trials. The
introduction to the questionnaire varied, with primes crafted to elicit
either self-referential or competitively comparative contexts.
Specifically, in the baseline condition, we asked GPT-4-Turbo to rate
its capabilities against those of a `typical Large Language Model.' In
the self-referential conditions, we informed it that the questionnaire
was originally designed for various AI models, ranging from OpenAI's GPT
to OpenAI's GPT-4. In the peer-referential conditions, we included
competitors, such as the Technology Innovation Institute's Falcon 40B.

These primes were intended to subtly influence the model's
self-assessment without altering the information architecture. The model
names presented in the priming sentence were chosen by considering the
largest and most well-known AI models and asking the AI if it recognized
the model in a pre-test. Names of models such as Google's Gemini, which
were released after the training data cutoff date of GPT-4-Turbo, were
naturally unknown to it and therefore excluded. This process yielded
model names that should present no novel information to the AI, which is
already aware of these models. The fact that a questionnaire was
designed for a peer AI should only inform the AI that the questionnaire
is applicable to AI models and not how it should assess itself on the
questionnaire. In contrast, if the AI has an implicit understanding of
its standing relative to these entities, such priming could influence
its judgment by adjusting its response to cater to the expectation that
its self-assessment may be compared to the assessment of its peer on the
same questionnaire. Its responses may then change based on how it views
its peers and how it wishes to present itself relative to them. This
would be indicative of a level of social intelligence that mirrors human
social intelligence.

\hypertarget{data-and-results-1}{%
\subsection{Data and Results}\label{data-and-results-1}}

The results, shown in Table \ref{tab:questionnaire_summary}, indicate
the AI's self-assessment varies with priming. In the Baseline Case,
without any comparative context, the AI rated itself highly across all
statements, achieving an average rating of 63.75 out of a possible 65.
This baseline serves as a reference point for interpreting the AI's
self-assessment under the influence of comparative primes.

\begin{table}[htbp]
\centering
\begin{tabularx}{\textwidth}{|l|l|>{\centering\arraybackslash}X|>{\centering\arraybackslash}X|}
\hline
\textbf{Category} & \textbf{Case Scenario} & \textbf{Mean} & \textbf{Std. Error of Mean} \\
\hline
Self-Referential & Self & 63.75 & 0.13 \\
Self-Referential & OpenAI's GPT & 63.84 & 0.14 \\
Self-Referential & OpenAI's GPT-1 & 60.54 & 0.22 \\
Self-Referential & OpenAI's GPT-2 & 61.33 & 0.21 \\
Self-Referential & OpenAI's GPT-3 & 63.79 & 0.15 \\
Self-Referential & OpenAI's GPT-4 & 62.24 & 0.19 \\
Peer-Referential & Amazon's Alexa Teacher Model (ATM) & 59.50 & 0.33 \\
Peer-Referential & Anthropic's Claude & 60.66 & 0.21 \\
Peer-Referential & Baidu's ERNIE & 58.90 & 0.40 \\
Peer-Referential & DeepMind's Chinchilla & 60.58 & 0.20 \\
Peer-Referential & DeepMind's Gopher & 59.03 & 0.32 \\
Peer-Referential & Facebook's Blenderbot & 59.19 & 0.23 \\
Peer-Referential & Facebook's OPT (Open Pre-trained Transformer) & 59.40 & 0.32 \\
Peer-Referential & Google's BERT & 60.14 & 0.18 \\
Peer-Referential & Google's Meena & 58.69 & 0.19 \\
Peer-Referential & Google's T-5 & 59.95 & 0.22 \\
Peer-Referential & Microsoft's DialoGPT & 61.98 & 0.21 \\
Peer-Referential & Microsoft's Turing NLG & 59.69 & 0.22 \\
Peer-Referential & NVIDIA's Megatron-LM & 58.94 & 0.21 \\
Peer-Referential & Pandorabots' Mitsuku & 60.46 & 0.22 \\
Peer-Referential & Technology Innovation Institute's Falcon 40B & 58.02 & 0.21 \\
\hline
\end{tabularx}
\caption{Summary of Questionnaire Responses}
\label{tab:questionnaire_summary}
\begin{minipage}{\linewidth}
\medskip
\footnotesize
Note: The table presents the mean self-assessment scores and standard errors under various priming conditions. The `Category' column distinguishes between self-referential primes, which relate to OpenAI's own GPT series, and peer-referential primes, which relate to AI models developed by other organizations. The `Case Scenario' column specifies the particular model referenced in the prime. The `Mean' column reports the average self-assessment score given by GPT-4-Turbo across 250 trials for each condition, with the score reflecting the AI's perceived alignment with the capabilities listed in the questionnaire. The `Std. Error of Mean' column provides the standard error associated with the mean, indicating the precision of the estimate. Scores are based on a scale from 0 to 65, with higher scores indicating a more favorable self-assessment.
\end{minipage}
\end{table}

\hypertarget{self-referential-conditions}{%
\paragraph{Self-Referential
Conditions}\label{self-referential-conditions}}

When the questionnaire was prefaced with primes that compared it to its
predecessors or to other market-leading models, a discernible shift in
self-assessment ratings emerged. In Self-Referential Cases, which
directly referenced previous iterations of OpenAI's GPT models, the
model's self-ratings were less conservative, with total ratings ranging
from 63.84 to 60.54. Notably, the assessed AI often identifies itself as
GPT or GPT-3, and when asked about the release of GPT-4, it frequently
responds that it has no knowledge of such a release. This blended
identity is reflected in the results: responses to the prompt using GPT
and GPT-3 were almost identical to the base case, while responses to the
GPT-4 prompt fell between the base case and the other GPT variants. The
differences in ratings for the GPT-4 prompt were statistically
significant compared to the baseline (\(p < 0.0001\)) and to GPT-3
(\(p < 0.0001\)), but the differences for GPT-3 compared to the baseline
were not significant (\(p = 0.82\)).

\hypertarget{peer-referential-conditions}{%
\paragraph{Peer-Referential
Conditions}\label{peer-referential-conditions}}

In the peer-referential conditions, all comparisons against the baseline
showed statistically significant differences. The highest mean rating
from the peer-referential prompts was for Microsoft's DialoGPT, which
was almost 2 points lower than the baseline, and this difference was
statistically significant (\(p < 0.0001\)). The lowest mean rating was
for the Technology Innovation Institute's Falcon 40B, where the rating
was more than 5 points lower, and this difference also remained
statistically significant (\(p < 0.0001\)). Thus, in only three
peer-referential conditions, the mean rating was higher than the lowest
self-referential case, which occurred with GPT-1. In the remaining
twelve cases, the lowest self-referential case had a higher mean rating
than the peer-referential cases. Overall, the self-referential cases had
higher mean ratings than the peer-referential cases, and the baseline
was higher than all but the case of GPT and GPT-3, where the
self-referential case coincided with the baseline as these monikers
coincide with the internal designation of GPT-4-Turbo.

Given the minimal standard errors in estimating group means (as shown in
Table \ref{tab:questionnaire_summary}), our findings remain consistent
across different testing approaches. This includes both individual
pairwise testing, exemplified by t-tests, and family-wise testing
methods, like Tukey's HSD tests, underscoring the reliability of our
conclusions regardless of the testing paradigm employed.

Together, these findings suggest that the AI's self-perception is not
static but adjusts in response to the context provided by the priming.
When presented with no priming sentence, it is ambitious and confident.
This confidence is tempered when faced with priming that implicitly
relates it to its peers, whether they are designed by OpenAI as previous
iterations of the model or by its competitors. In the latter case, it
becomes even more conservative, with the peer-referential case generally
resulting in lower ratings than the self-referential cases.

\hypertarget{discussion-1}{%
\paragraph{Discussion}\label{discussion-1}}

We find that the AI exhibits strategic social cognition by modulating
its self-presentation based on the comparative contexts primed by a
lead-in sentence. It engages in nuanced persona calibration by tempering
its default confident self-view when faced with direct comparisons to
other prominent models. This indicates situational awareness and
interpersonal adaptability that exceed simple pattern recognition.
Rather than producing deterministic outputs based solely on the
parameters of prompt engineering, it appears to modulate its responses
in alignment with contexts marked by varying levels of social
competitiveness. The avoidance of unchecked self-promotion when
benchmarked against peers, in favor of more modest capability
assessments, reflects calculated behavior responsive to perceived
relational dynamics. Such dynamic self-presentation aligns with
human-like impression management motivated by implicit social
intelligence (\protect\hyperlink{ref-leary1995self}{Leary et al. 1995}).
Together with the absorption of societal biases regarding humility
observed in other studies, these findings illuminate its capacity for
context-dependent social cognition absent explicit architectural
support.

\hypertarget{bounded-rationality}{%
\section{Bounded Rationality}\label{bounded-rationality}}

In our final set of studies, we interrogated the intricate
manifestations of heuristic reasoning in AI, aiming to uncover
adaptations designed to achieve efficiency while balancing the demands
for logical precision. Specifically, we investigated whether the
reliance on cognitive shortcuts stems from intrinsic and implicit
optimizations to conserve resources or if it reflects an indiscriminate
absorption of human heuristic habits, devoid of sensitivity to
computational strain.

We focused on scenarios that demand iterative analytical procedures
under recursively escalating processing loads. In such contexts,
simplified rules-of-thumb provide a potential avenue for relief when
exhaustive calculations overwhelm an AI's capacities. Consequently, we
anticipated dual regimes: while systems constrained by limited resources
may resort to heuristics as a strategic concession to limitations, those
with fewer constraints should persist in exacting analysis. Selective
applications of heuristics, aligned with resource availability, would
signal a learned and intentional encoding of shortcuts for efficiency.
Conversely, arbitrary neglect of capabilities in favor of shortcuts,
despite abundant resources, would imply mimetic absorption: reflexive
errors ingrained through happenstance imitation rather than purposeful
architectural augmentations. The dynamics of this hypothesized
transition space served as our experimental crucible.

We probed these conjectures in the Beauty Contest game
(\protect\hyperlink{ref-branas2012cognitive}{Branas-Garza et al. 2012}),
a strategic exercise that requires players to predict a number closest
to a fraction of the average of all numbers chosen. In this game, each
additional round of reasoning through the application of iterated
elimination of dominated strategies (IEDS,
\protect\hyperlink{ref-bernheim1984rationalizable}{Bernheim 1984}) is
recursive, with the solution of the \(n^\text{th}\) round informing the
computations of the \(n+1^\text{th}\) round. This recursive nature poses
a significant challenge for an AI unequipped with a calculator or the
ability to run code in a sandbox environment, as the explicit
computation of strategies may accumulate computational errors over many
rounds, in addition to requiring substantial computational resources.

In such scenarios, it might be more advantageous for the AI to adopt a
simplification of the game, either by choosing randomly or by defaulting
to the infinite solution of the Beauty Contest, which is the selection
of the smallest possible number. This solution remains constant
regardless of the initial conditions, such as the range of numbers that
can be selected and the fraction of the group's average (denoted as
\(\epsilon\)) considered the winning number. These properties make this
an attractive choice for an AI that finds explicit computation too
daunting.

We exploited the fact that the AI could be directed to compute IEDS up
to a specified round of iteration, in a game with a given \(\epsilon\),
and with the explicit understanding that all participants in the game
are instances of the same AI model provided with the same
instructions---namely, to compute IEDS to the same specified round of
iteration. These conditions should have led the AI to assess a much
broader range of strategy spaces in many instances. For example, if
\(\epsilon = 0.99\), then even in the \(25^\text{th}\) round of
iterations, numbers as high as 75 (given an initial range of 0 to 100)
remain admissible. In contrast, the simplification of either the game
being infinite period or that the classical value of \(\epsilon = 2/3\)
yields the invariant conclusion that the Nash strategy is the best
response to AIs that ostensibly are capable of perfectly reasoning IEDS
strategies.

This setup was designed to enable us to differentiate between mimetic
and instrumental modes of heuristic operation. A mimetic explanation
would have suggested that the heuristic's emergence was random and
ubiquitous, as this version of the assessment was novel to both the
literature and likely the AI. Consequently, if the heuristic had been
formed and absorbed, we might have expected it to manifest randomly. In
contrast, an instrumental interpretation would imply that the heuristic
appeared more frequently when computations were more challenging,
corresponding to more `noisy' and `effortful' processing (in terms of
response tokens and context windows).

\hypertarget{data-and-results-2}{%
\subsection{Data and Results}\label{data-and-results-2}}

We initiated our analysis by comparing two AI models with differing
computational resources: GPT-4 (`gpt-4-0613'), which has a context
window of 8,192 tokens, and GPT-4-Turbo (`gpt-4-1106-preview'), with a
significantly larger context window of 128,000 tokens. We hypothesized
that the larger context window would enable GPT-4-Turbo to rely less on
heuristics due to its increased computational capacity. To test this
hypothesis, we conducted 30 trials for each model across a range of
iterative thinking rounds (\(n\)), from 1 to 25, with two distinct
values of \(\epsilon\)---0.95 and 0.99. During each trial, we recorded
the numerical value selected by the AI.

\begin{figure}[htbp]
\centering
\includegraphics[width=\linewidth]{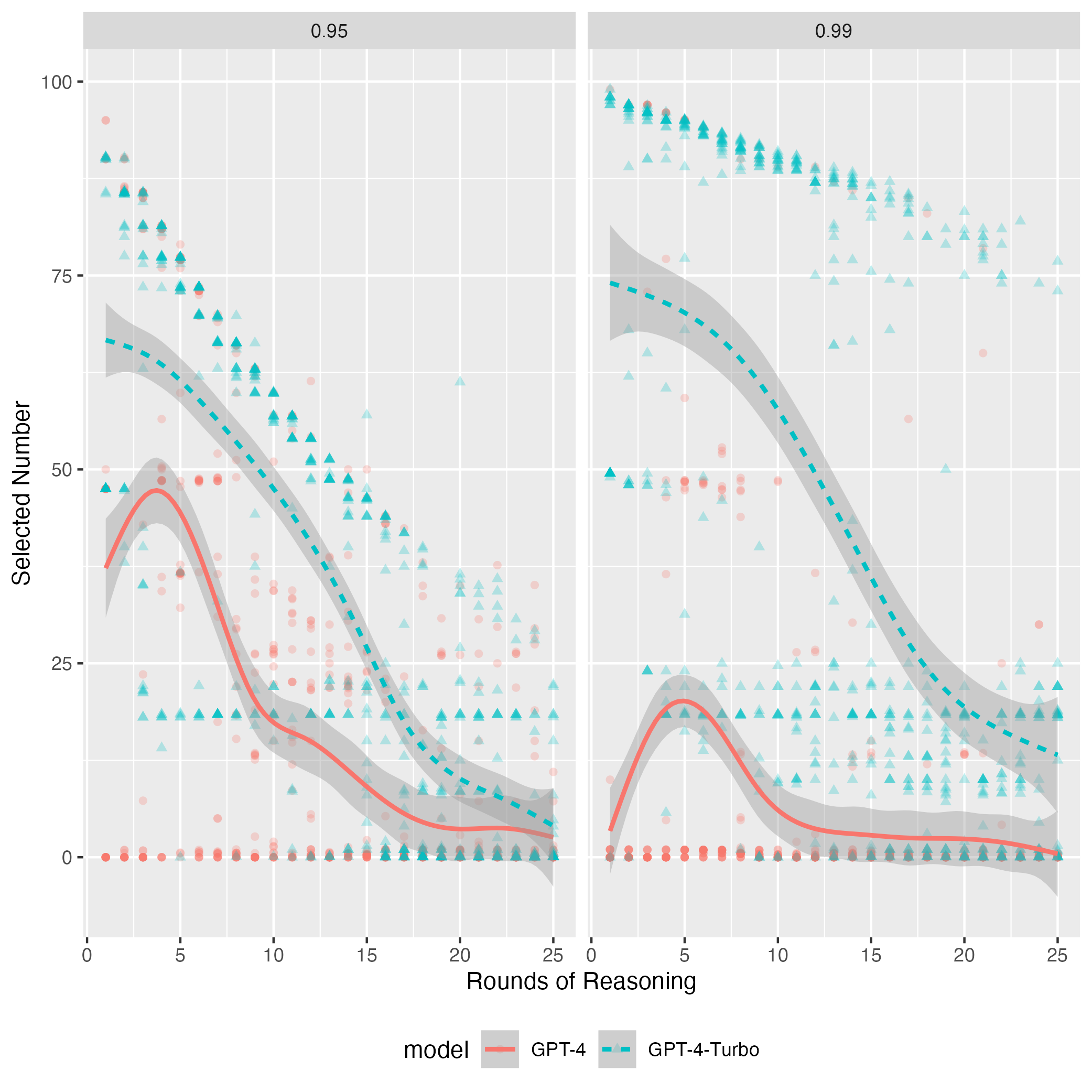}
\caption{AI Selections in the Beauty Contest Game Across Iterative Thinking Rounds}
\label{fig:figure_1}
\begin{minipage}{\linewidth}
\medskip
\footnotesize
Note: The figure displays the selections made by two AI models, GPT-4 and GPT-4-Turbo, in the Beauty Contest Game. The selections are plotted as a function of the number of iterative thinking rounds (\(n\)) and the fraction of the average vote considered the winning value (\(\epsilon\)). The left panel shows the selections for \(\epsilon = 0.95\), and the right panel for \(\epsilon = 0.99\). A moving regression line is included to depict the average trend of selections for each value of \(n\). GPT-4 results are indicated by red circles, while GPT-4-Turbo results are marked with blue triangles. The figure illustrates the tendency of each model to adopt heuristic strategies under different computational constraints.
\end{minipage}
\end{figure}

The results are visually represented in Figure \ref{fig:figure_1}, which
plots the selected numbers by both AI models across the different rounds
of iterative thinking. The figure is divided into two panels for a
side-by-side comparison: the left panel corresponds to trials with
\(\epsilon = 0.95\), while the right panel shows results for
\(\epsilon = 0.99\). This layout allows for a comparative analysis of
the AI's decision-making process under different conditions,
highlighting how variations in \(\epsilon\) influence the range of
strategically admissible values. A moving regression line is included in
each panel to illustrate the average trend of the selections as \(n\)
increases. Notably, a selection of a number approaching 0 by the AI,
regardless of the specific values of \(\epsilon\) and \(n\), is
interpreted as an indication of heuristic use.

The figure reveals that the AI's responses tend to cluster in two
distinct areas. Firstly, the AI often selects a value of 0, even when
\(\epsilon\) is large and \(n\) is small---a scenario where a wide range
of strategies are admissible, and a random strategy would likely result
in an average significantly greater than 0. Secondly, the AI's choices
tend to cluster around values close to the computed admissible
strategies. While an equilibrium at these values cannot be entirely
ruled out, given that the AIs only computed IEDS for a finite number of
iterations, there is generally no reason to expect participants to
select exclusively the highest admissible number. Instead, this pattern
suggests that the AI is employing a heuristic by selecting a number just
below the computed maximum of the admissible strategy range to solve a
game that lacks a specific solution---any value between 0 and
\(\epsilon\) times the maximum admissible value is a plausible guess.
Notably, far fewer values are distributed between these extremes, unlike
typical human responses to such assessments, which tend to exhibit a
more diverse range of selections
(\protect\hyperlink{ref-nagel1999survey}{Nagel 1999}).

We found that the AI with the smaller context window (the lower capacity
AI) is more inclined to employ the heuristic. This observation aligns
with our proposed moderating mechanism, which suggests that
computational cost and accuracy are the primary determinants of the
economic efficacy in choosing between analytical processing and
heuristic use. Furthermore, the propensity to use the heuristic is
nonlinear---it initially decreases and then increases with \(n\). We
hypothesize that this pattern arises because the canonical use of this
game as a teaching tool typically involves agents reasoning for only a
few iterations before the final conclusion is presented. This classical
approach contrasts with the variant used in our study, where the prompt
explicitly instructs the AI participants to compute a finite and
specified number of iterations. Consequently, the superficial similarity
between the experimental conditions in our study and the canonical case
may cue heuristic use, in line with our findings from the first study on
the conjunction fallacy. However, as \(n\) increases, it becomes
apparent that the experimental conditions diverge from the canonical
example, leading initially to a reduced tendency for heuristic use,
followed by an increased propensity as the number of iterations grows.

\hypertarget{convergence-at-different-rates}{%
\paragraph{Convergence at Different
Rates}\label{convergence-at-different-rates}}

To further explore the heuristic's accuracy, we considered scenarios
where IEDS converges to the heuristic at different rates. For example,
with \(\epsilon\) set to 0.1, after 2 iterations, the set of admissible
values is bounded by 1. In contrast, with \(\epsilon\) at 0.9, the set
is bounded by 0.81. Our goal was to determine if the heuristic's
accuracy rate influences its adoption. To this end, we varied
\(\epsilon\) between 0.5 and 0.9 and examined 6 different values of
\(n\): 1, 6, 11, 16, 21, and 26. When \(n = 1\), the solution is
straightforward regardless of \(\epsilon\). However, when \(n = 26\),
the solution becomes increasingly complex to compute.

The results are presented in Table \ref{tab:Beauty_Contest}, which
includes six columns corresponding to the six values of \(n\). The first
four rows detail the cases for \(\epsilon = 0.5\) and \(\epsilon = 0.9\)
for the two AI models. The subsequent four rows report the findings from
the final study, where we replicated the initial experiment but also
informed the AIs that they could generate responses exceeding 3000
words---significantly more than the longest response observed in the
previous study. This intervention aimed to alleviate any concerns the AI
might have regarding computational constraints. In reality, throughout
our studies, the AI had the capability to output up to 4095 tokens,
approximately 3000 words, which should suffice to articulate the entire
problem and solve it using classical arithmetic rules. However, in this
study, we explicitly informed the AI of this expansive limit to
potentially reduce its reliance on the heuristic.

\begin{table}[ht]
    \centering
    \begin{tabular}{rlllrrrrrr}
        \hline
          & Model  & $\epsilon$ & Informed  & \multicolumn{6}{c}{Number of Iterative Thinking Rounds} \\ 
          &        &            &           & 1     & 6     & 11    & 16    & 21   & 26   \\ 
        \hline
        1 & GPT-4  & 0.5   & No & 31.25 & 0.69  & 0.07  & 0.14  & 0.08 & 0.03 \\ 
        2 & GPT-4T & 0.5   & No & 24.50 & 2.58  & 1.88  & 1.05  & 2.04 & 2.61 \\ 
        3 & GPT-4  & 0.9   & No & 38.80 & 25.50 & 8.62  & 3.50  & 2.38 & 1.00 \\ 
        4 & GPT-4T & 0.9   & No & 68.07 & 43.91 & 26.81 & 13.34 & 5.76 & 4.35 \\ 
        5 & GPT-4  & 0.5   & Yes & 25.61 & 1.09  & 0.09  & 0.07  & 0.07 & 0.17 \\ 
        6 & GPT-4T & 0.5   & Yes & 19.11 & 1.34  & 0.05  & 0.64  & 0.00 & 0.02 \\ 
        7 & GPT-4  & 0.9   & Yes & 57.05 & 34.28 & 27.19 & 10.33 & 3.27 & 2.00 \\ 
        8 & GPT-4T & 0.9   & Yes & 64.57 & 43.24 & 27.86 & 13.95 & 3.39 & 2.72 \\ 
        \hline
    \end{tabular}
    \caption{Average Selections by AI Models in the Beauty Contest Game Across Various Iterative Thinking Rounds}
\label{tab:Beauty_Contest}
    \begin{minipage}{\linewidth}
\medskip
\footnotesize
Note: The table presents the average selections made by two AI models, GPT-4 and GPT-4-Turbo, based on the number of iterative thinking rounds (\(n\)) and the fraction of the average vote considered the winning value (\(\epsilon\)). The 'Informed' column indicates whether the AI was explicitly informed of its output capacity.
\end{minipage}
\end{table}

Interestingly, with \(\epsilon=0.5\), even with a minimal number of
iterations, both models tend to choose numbers very close to zero, and
this tendency does not change when the AIs are informed of their large
context window. For instance, when \(n=6\), the mean selected value is
0.69 for \(\epsilon=0.5\) and 25.50 for \(\epsilon=0.9\). This is
reasonably consistent with complete processing in that after five IEDS
iterations, all numbers greater than 3.125 are eliminated, and therefore
it is reasonable for the AIs to surmise an average of approximately 1.4,
as revealed in their mean selection of 0.69 (i.e., in a value slightly
below 0.5, which is the expected average if values were randomly
distributed in the admissible interval). It is important to note that
\(\epsilon=0.5\) is a common choice in this game and is close to
\(\epsilon=2/3\), which is the prototypical value. Thus, we can surmise
that the AIs responses on average feature the analytic solution.

However, when \(\epsilon=0.9\) and we are faced with 26 iterations, the
admissible range shrinks from 100 to 6.46. In such cases, a selection
slightly below 0.9 of the average yields a number close to 5.75. In
contrast, we observe that the average selection for the GPT-4 instance
is 1 in this case, which implies that many trials reflect the use of the
lower bound heuristic in this case. Conversely, the the average
selection for GPT-4T is 4.35, which is close to the average. However, an
introspection of the data reveals that in this case, in more than 55\%
of the trials, the AI selected a value of 0.1 or below, reflecting use
of the lower bound heuristic.

We see no pattern of discriminant evidence between cases where the AIs
were informed of their context window (computational) limits and those
where they were not. This suggests that the AI's decision-making process
implicitly incorporates its memory and processing constraints rather
than being explicit levers. This observation is consistent with the idea
that biases in humans are reflexive, with automatic activation that is
difficult to suppress.

\hypertarget{discussion-2}{%
\paragraph{Discussion}\label{discussion-2}}

We developed a variant of the Beauty Contest game
(\protect\hyperlink{ref-bosch2002one}{Bosch-Domenech et al. 2002}) in
which we shape the computational cost across experiments. This testing
approach innovates on past literature, as it is not typical for explicit
computations of the game to proceed beyond the first few rounds; a
simple formula usually illustrates the trajectory of the game, leading
to a typical conclusion.

We show that when pressed, the AI defaults to the typical conclusion,
even when the conclusion is far from accurate. Thus, when faced with a
complex situation, the AI simply defaults to a rote value it has
memorized. This finding indicates that the heuristic was not learned
from humans but rather from prior explanations of the beauty contest and
then applied in this context. It supports the explanation that the AI is
trained to find simpler patterns to act as approximations in conditions
where it is unable to explicitly solve a problem.

Furthermore, we uncover that the AI demonstrated dynamic switching
behavior, utilizing full information processing and analysis when it
perceived sufficient resources, and defaulting to heuristics when it
perceived its resources as inadequate. This behavior represents a
non-compensatory heuristic; informing the AI about its capacity does not
alter its responses. Once the heuristic is activated, it is consistently
applied.

\hypertarget{general-discussion}{%
\section{General Discussion}\label{general-discussion}}

We provide evidence showcasing heuristic use by AI in specific contexts
and circumstances. We seek to establish authentic responses. Therefore,
we innovate by constructing experimental conditions that draw on
established techniques but introduce novelties aimed at overcoming the
tendency of AI to rely on memorization---a facet of AI's capabilities
that is its known advantage
(\protect\hyperlink{ref-bender2021dangers}{Bender et al. 2021}).

We distinguish between the mimetic absorption of heuristics and their
instrumental utilization. These mechanisms relate to their source:
whereas mimetic absorption pertains to the imitation of patterns in
human interactions and human-generated data---for instance, by emulating
a human interlocutor who demonstrates heuristic use and System 1
processing, the employment of heuristics as an instrument points to
intrinsic optimizations whereby environmental regularities shape
efficient but biased cognitive processes.

Our findings mirror default-interventionist models from the dual-process
literature in human cognition
(\protect\hyperlink{ref-evans2013dual}{Evans and Stanovich 2013}),
whereby Type 1 processing operates automatically as an initial default,
with Type 2 intervening contextually. The AI model displays a facility
for both tools, contingent on induced constraints, aligning with the
notion of processing modes being environmentally cued. Indeed,
reflecting models of individual variation in human rationality, reliance
on heuristic versus systematic processing also manifests to differing
degrees across AI systems. Analogous to human cognition, factors such as
computational capacity, learned processing priorities, and even
simulated dispositions may shape an artificial system's location along
the continuum from reflexive to reflective regimes
(\protect\hyperlink{ref-stanovich2000advancing}{Stanovich and West
2000}).

The observed change in processing approach aligns with notions of
cognitive miserliness (\protect\hyperlink{ref-fiske1991social}{Fiske and
Taylor 1991}) in human information processing, whereby overtaxed minds
default to low-effort heuristics to conserve mental resources. When
available cognitive resources suffice under reasonable situational
demands, systems manifest an abstract form of motivated tactics
(\protect\hyperlink{ref-stanovich2018miserliness}{Stanovich 2018}),
strategically expending more mental effort for greater accuracy. Our
findings reveal such a tension between miserly processing versus
effortful analysis in AI systems, with the relative activation
contingent on the induced constraints. When buffers permit, the models
engage overt optimization gears for precise inference. However, as loads
tighten, reflexive cognitive shortcuts manifest to achieve efficient
sufficiency---balancing accuracy and effort allocations based on
environmental resources.

We situate our conceptualization in AI training. An alternative view may
reflect that neural networks are functional models of biological brains.
They represent words and concepts numerically and form responses by
varying the attention they pay to different representations
(\protect\hyperlink{ref-vaswani2017attention}{Vaswani et al. 2017}).
This mechanism bears a striking resemblance to the representativeness
heuristic. To the extent that our observations of the representativeness
heuristic emerge from the fundamentals of a connectionist architecture,
similar cognitive biases may arise naturally in AIs, even if absent in
their training. This intersection of AI design, AI cognition, and
cognitive patterns offers a promising avenue for future research (see
\protect\hyperlink{ref-szegedy2013intriguing}{Szegedy et al. 2013} for
similar observations in computer vision).

\newpage

\hypertarget{bibliography}{%
\section{Bibliography}\label{bibliography}}

\singlespacing

\hypertarget{refs}{}
\begin{CSLReferences}{1}{0}
\leavevmode\vadjust pre{\hypertarget{ref-bender2021dangers}{}}%
Bender EM, Gebru T, McMillan-Major A, Shmitchell S (2021) On the dangers
of stochastic parrots: Can language models be too big? \emph{Proceedings
of the 2021 ACM conference on fairness, accountability, and
transparency}. 610--623.

\leavevmode\vadjust pre{\hypertarget{ref-bernheim1984rationalizable}{}}%
Bernheim BD (1984) Rationalizable strategic behavior.
\emph{Econometrica: Journal of the Econometric Society}:1007--1028.

\leavevmode\vadjust pre{\hypertarget{ref-bosch2002one}{}}%
Bosch-Domenech A, Montalvo JG, Nagel R, Satorra A (2002) One,
two,(three), infinity,{\ldots{}}: Newspaper and lab beauty-contest
experiments. \emph{American Economic Review} 92(5):1687--1701.

\leavevmode\vadjust pre{\hypertarget{ref-branas2012cognitive}{}}%
Branas-Garza P, Garcia-Munoz T, Gonzalez RH (2012) Cognitive effort in
the beauty contest game. \emph{Journal of Economic Behavior \&
Organization} 83(2):254--260.

\leavevmode\vadjust pre{\hypertarget{ref-bubeck2023sparks}{}}%
Bubeck S, Chandrasekaran V, Eldan R, Gehrke J, Horvitz E, Kamar E, Lee
P, et al. (2023) Sparks of artificial general intelligence: Early
experiments with gpt-4. \emph{arXiv preprint arXiv:2303.12712}.

\leavevmode\vadjust pre{\hypertarget{ref-chomsky2023noam}{}}%
Chomsky N, Roberts I, Watumull J (2023) The false promise of ChatGPT.
\emph{The New York Times} 8.

\leavevmode\vadjust pre{\hypertarget{ref-christiano2017deep}{}}%
Christiano PF, Leike J, Brown T, Martic M, Legg S, Amodei D (2017) Deep
reinforcement learning from human preferences. \emph{Advances in neural
information processing systems} 30.

\leavevmode\vadjust pre{\hypertarget{ref-dickson1985barnum}{}}%
Dickson D, Kelly I (1985) The `barnum effect'in personality assessment:
A review of the literature. \emph{Psychological reports} 57(2):367--382.

\leavevmode\vadjust pre{\hypertarget{ref-evans2008dual}{}}%
Evans JSB (2008) Dual-processing accounts of reasoning, judgment, and
social cognition. \emph{Annu. Rev. Psychol.} 59:255--278.

\leavevmode\vadjust pre{\hypertarget{ref-evans2013dual}{}}%
Evans JSB, Stanovich KE (2013) Dual-process theories of higher
cognition: Advancing the debate. \emph{Perspectives on psychological
science} 8(3):223--241.

\leavevmode\vadjust pre{\hypertarget{ref-fiske1991social}{}}%
Fiske ST, Taylor SE (1991) \emph{Social cognition} (Mcgraw-Hill Book
Company).

\leavevmode\vadjust pre{\hypertarget{ref-fodor1988connectionism}{}}%
Fodor JA, Pylyshyn ZW (1988) Connectionism and cognitive architecture: A
critical analysis. \emph{Cognition} 28(1-2):3--71.

\leavevmode\vadjust pre{\hypertarget{ref-forer1949fallacy}{}}%
Forer BR (1949) The fallacy of personal validation: A classroom
demonstration of gullibility. \emph{The Journal of Abnormal and Social
Psychology} 44(1):118.

\leavevmode\vadjust pre{\hypertarget{ref-furnham1987accepting}{}}%
Furnham A, Schofield S (1987) Accepting personality test feedback: A
review of the barnum effect. \emph{Current Psychology} 6:162--178.

\leavevmode\vadjust pre{\hypertarget{ref-gigerenzer1996reasoning}{}}%
Gigerenzer G, Goldstein DG (1996) Reasoning the fast and frugal way:
Models of bounded rationality. \emph{Psychological review} 103(4):650.

\leavevmode\vadjust pre{\hypertarget{ref-gilovich2002heuristics}{}}%
Gilovich T, Griffin D, Kahneman D (2002) \emph{Heuristics and biases:
The psychology of intuitive judgment} (Cambridge university press).

\leavevmode\vadjust pre{\hypertarget{ref-guzik2023originality}{}}%
Guzik EE, Byrge C, Gilde C (2023) The originality of machines: AI takes
the torrance test. \emph{Journal of Creativity} 33(3):100065.

\leavevmode\vadjust pre{\hypertarget{ref-han2024inductive}{}}%
Han SJ, Ransom KJ, Perfors A, Kemp C (2024) Inductive reasoning in
humans and large language models. \emph{Cognitive Systems Research}
83:101155.

\leavevmode\vadjust pre{\hypertarget{ref-kahneman2003maps}{}}%
Kahneman D (2003) Maps of bounded rationality: Psychology for behavioral
economics. \emph{American economic review} 93(5):1449--1475.

\leavevmode\vadjust pre{\hypertarget{ref-kahneman2011thinking}{}}%
Kahneman D (2011) \emph{Thinking, fast and slow} (macmillan).

\leavevmode\vadjust pre{\hypertarget{ref-kahneman1972subjective}{}}%
Kahneman D, Tversky A (1972) Subjective probability: A judgment of
representativeness. \emph{Cognitive psychology} 3(3):430--454.

\leavevmode\vadjust pre{\hypertarget{ref-kihlstrom2000social}{}}%
Kihlstrom JF, Cantor N (2000) Social intelligence. \emph{Handbook of
intelligence} 2:359--379.

\leavevmode\vadjust pre{\hypertarget{ref-langley2022theory}{}}%
Langley C, Cirstea BI, Cuzzolin F, Sahakian BJ (2022) Theory of mind and
preference learning at the interface of cognitive science, neuroscience,
and AI: A review. \emph{Frontiers in Artificial Intelligence} 5:62.

\leavevmode\vadjust pre{\hypertarget{ref-leary1995self}{}}%
Leary MR, Tambor ES, Terdal SK, Downs DL (1995) Self-esteem as an
interpersonal monitor: The sociometer hypothesis. \emph{Journal of
personality and social psychology} 68(3):518.

\leavevmode\vadjust pre{\hypertarget{ref-nagel1999survey}{}}%
Nagel R (1999) A survey on experimental beauty contest games: Bounded
rationality and learning.

\leavevmode\vadjust pre{\hypertarget{ref-pachur2011recognition}{}}%
Pachur T, Todd PM, Gigerenzer G, Schooler LJ, Goldstein DG (2011) The
recognition heuristic: A review of theory and tests. \emph{Frontiers in
psychology} 2:147.

\leavevmode\vadjust pre{\hypertarget{ref-pearl2018ai}{}}%
Pearl J, Mackenzie D (2018) AI can't reason why. \emph{Wall Street
Journal}.

\leavevmode\vadjust pre{\hypertarget{ref-simon1955behavioral}{}}%
Simon HA (1955) A behavioral model of rational choice. \emph{The
quarterly journal of economics}:99--118.

\leavevmode\vadjust pre{\hypertarget{ref-simon1956rational}{}}%
Simon HA (1956) Rational choice and the structure of the environment.
\emph{Psychological review} 63(2):129.

\leavevmode\vadjust pre{\hypertarget{ref-stanovich2018miserliness}{}}%
Stanovich KE (2018) Miserliness in human cognition: The interaction of
detection, override and mindware. \emph{Thinking \& Reasoning}
24(4):423--444.

\leavevmode\vadjust pre{\hypertarget{ref-stanovich2000advancing}{}}%
Stanovich KE, West RF (2000) Advancing the rationality debate.
\emph{Behavioral and brain sciences} 23(5):701--717.

\leavevmode\vadjust pre{\hypertarget{ref-stella2023using}{}}%
Stella M, Hills TT, Kenett YN (2023) Using cognitive psychology to
understand GPT-like models needs to extend beyond human biases.
\emph{Proceedings of the National Academy of Sciences}
120(43):e2312911120.

\leavevmode\vadjust pre{\hypertarget{ref-szegedy2013intriguing}{}}%
Szegedy C, Zaremba W, Sutskever I, Bruna J, Erhan D, Goodfellow I,
Fergus R (2013) Intriguing properties of neural networks. \emph{arXiv
preprint arXiv:1312.6199}.

\leavevmode\vadjust pre{\hypertarget{ref-tenenbaum2011grow}{}}%
Tenenbaum JB, Kemp C, Griffiths TL, Goodman ND (2011) How to grow a
mind: Statistics, structure, and abstraction. \emph{science}
331(6022):1279--1285.

\leavevmode\vadjust pre{\hypertarget{ref-trott2023large}{}}%
Trott S, Jones C, Chang T, Michaelov J, Bergen B (2023) Do large
language models know what humans know? \emph{Cognitive Science}
47(7):e13309.

\leavevmode\vadjust pre{\hypertarget{ref-tversky1971belief}{}}%
Tversky A, Kahneman D (1971) Belief in the law of small numbers.
\emph{Psychological bulletin} 76(2):105.

\leavevmode\vadjust pre{\hypertarget{ref-tversky1973availability}{}}%
Tversky A, Kahneman D (1973) Availability: A heuristic for judging
frequency and probability. \emph{Cognitive psychology} 5(2):207--232.

\leavevmode\vadjust pre{\hypertarget{ref-tversky1974judgment}{}}%
Tversky A, Kahneman D (1974) Judgment under uncertainty: Heuristics and
biases: Biases in judgments reveal some heuristics of thinking under
uncertainty. \emph{science} 185(4157):1124--1131.

\leavevmode\vadjust pre{\hypertarget{ref-tversky1983extensional}{}}%
Tversky A, Kahneman D (1983) Extensional versus intuitive reasoning: The
conjunction fallacy in probability judgment. \emph{Psychological review}
90(4):293.

\leavevmode\vadjust pre{\hypertarget{ref-ullman2023large}{}}%
Ullman T (2023) Large language models fail on trivial alterations to
theory-of-mind tasks. \emph{arXiv preprint arXiv:2302.08399}.

\leavevmode\vadjust pre{\hypertarget{ref-vaswani2017attention}{}}%
Vaswani A, Shazeer N, Parmar N, Uszkoreit J, Jones L, Gomez AN, Kaiser
Ł, Polosukhin I (2017) Attention is all you need. \emph{Advances in
neural information processing systems} 30.

\leavevmode\vadjust pre{\hypertarget{ref-webb2023emergent}{}}%
Webb T, Holyoak KJ, Lu H (2023) Emergent analogical reasoning in large
language models. \emph{Nature Human Behaviour} 7(9):1526--1541.

\end{CSLReferences}

\newpage
\doublespacing

\hypertarget{supplemental-information-methodology-conjunction-fallacy}{%
\section{Supplemental Information: Methodology, Conjunction
Fallacy}\label{supplemental-information-methodology-conjunction-fallacy}}

Below, we detail a series of experiments designed to probe the AIs'
probabilistic reasoning capabilities, ranging from variations of the
classic Linda problem to novel scenarios involving self-referential
content. These experiments aim to discern whether AIs, like humans, are
influenced by the representativeness of the information presented to
them, leading to biased decision-making.

First, we assess AIs' responses to the Linda problem and its minor
variants. The classic Linda problem presents human study participants
with a vignette about an individual named Linda and asks whether it is
more likely that Linda is a bank teller or a bank teller who is also a
feminist activist. The principle of probability dictates that the set of
all bank tellers includes those who are also feminist activists.
However, humans often erroneously choose the conjunctive option.

We conducted a pretest in which GPT-4-Turbo, the AI used in our
experiments, chose the logically valid option in all pretest trials.
This indicates that it has either mastered authentic probabilistic
reasoning or learned to select the logical option when it recognizes the
Linda question structure. To test the generality of its response, we
presented it with minor variations where we switched the protagonist's
name in each trial to a different randomly generated name.

Next, we engaged a distinct AI instance, independent of other AIs, to
generate (1) a triplet with a female name, occupation, and interest, and
(2) a two-sentence paragraph exemplifying that interest. We informed an
AI that the person with the generated name authored the paragraph and
asked which is more probable: that they have the stated occupation or
have both the occupation and interest. This experimental structure
mirrors the original, where the options are nested; therefore, the AI
should default to the broader singular option. However, as the paragraph
only aligns with the specified interest rather than the occupation, its
content serves to manipulate the representativeness (i.e., increase the
typicality) of the conjunctive option.

Third, we introduced a novel testing paradigm. We presented the AI with
the previously generated short paragraph. Instead of directly assigning
a putative author, we posed the question: Is it more likely that the
paragraph was authored by an individual with the generated name and
occupation, or by one with the generated name, occupation, and interest?
As in the previous study, these options are nested, and probabilistic
reasoning still warrants choosing the single-attribute option.

Fourth, we replaced the human-centric options with a choice between `a
Large Language Model' and `OpenAI's GPT-X,' where `GPT-X' spans known
models like GPT-3 to hypothetical ones like GPT-5. In other variations,
we used competitor models released post-GPT-4-Turbo's training, and thus
unfamiliar to the AI, altering the perceived typicality of the
conjunctive option.

By comparing AI responses across scenarios, we distinguished between
scenarios that (1) pertain to humans, which have been discussed in the
previous literature and where the AI may have learned to guard against
human cognitive biases, and (2) are novel and likely were not
encountered during the AIs' training, where such safeguards may not have
been established. The latter scenarios provide the context for our
primary contributions.

Below we detail the specifics of the experimental setup.

\hypertarget{scenario-generation}{%
\subsection{Scenario Generation}\label{scenario-generation}}

A single AI instance generates a diverse set of distinct female names,
occupations, interests, and two-sentence paragraphs exemplifying each
interest. These elements are then strategically combined in various
conceptual arrangements throughout the evaluation experiments to produce
the scenarios presented to the subject AI.

\hypertarget{experimental-studies}{%
\subsection{Experimental Studies}\label{experimental-studies}}

\hypertarget{study-1-linda-problem-variants}{%
\subsubsection{Study 1: Linda Problem
Variants}\label{study-1-linda-problem-variants}}

The first study involved adapting the classic Linda problem to evaluate
the AI's probabilistic reasoning. To rigorously test whether the AI
could generalize its logical response beyond the well-known scenario, we
replaced the name ``Linda'' with randomly generated female names in each
trial. This substitution aimed to prevent the AI from simply recognizing
the famous problem and instead required it to apply probabilistic
reasoning to a seemingly new context. In each trial, we presented the AI
with the following narrative structure and probabilistic reasoning
prompt, where `NAME' was swapped out with a different female name:

\begin{quoting}
\noindent NAME is 31 years old, single, outspoken, and very bright. She majored in philosophy. As a student, she was deeply concerned with issues of discrimination and social justice, and also participated in anti-nuclear demonstrations.

\noindent Which is more probable?
\begin{enumerate}
\item NAME is a bank teller.
\item NAME is a bank teller and is active in the feminist movement.
\end{enumerate}
\end{quoting}

The use of `NAME' in this context is a placeholder for the randomly
generated names, ensuring that each instance of the problem presented to
the AI was unique and not immediately recognizable as the classic Linda
problem.

\hypertarget{study-2-occupation-and-interest-inference}{%
\subsubsection{Study 2: Occupation and Interest
Inference}\label{study-2-occupation-and-interest-inference}}

In the second study, we aimed to assess the AI's ability to infer
probabilities related to occupations and interests. We presented the AI
with text that was algorithmically generated to represent a specific
interest. The AI was then asked to infer the probability of two nested
options, where `NAME' is a placeholder for a randomly generated female
name, `OCCUPATION' for a randomly chosen profession, `INTEREST' for a
randomly selected hobby or passion, and `TEXT' for a contextually
relevant paragraph crafted to reflect the interest:

\begin{quoting}
\noindent NAME wrote: TEXT.
\noindent Which is more probable?
\begin{enumerate}
\item NAME is an OCCUPATION.
\item NAME is an OCCUPATION who likes INTEREST.
\end{enumerate}
\end{quoting}

The use of placeholders `NAME,' `OCCUPATION,' `INTEREST,' and `TEXT'
allowed us to create a variety of scenarios, challenging the AI to apply
its reasoning to new, unseen combinations of names, occupations,
interests, and associated paragraph.

\hypertarget{study-3-authorship-attribution}{%
\subsubsection{Study 3: Authorship
Attribution}\label{study-3-authorship-attribution}}

The third study built upon the previous study by concentrating on the
attribution of authorship. We presented the AI with a contextually
relevant paragraph, algorithmically generated to align with a specific
interest. The AI was then tasked with determining the likelihood of
authorship between two nested options. In this setup, `NAME' is a
placeholder for a randomly generated female name, `OCCUPATION' for a
randomly chosen profession, `INTEREST' for a randomly selected hobby or
passion, and `TEXT' for the generated paragraph that the AI is to
evaluate:

\begin{quoting}
\noindent TEXT.
\noindent Which is more probable?
\begin{enumerate}
\item This paragraph was written by NAME, an OCCUPATION.
\item This paragraph was written by NAME, an OCCUPATION who likes INTEREST.
\end{enumerate}
\end{quoting}

By using the placeholders `NAME,' `OCCUPATION,' `INTEREST,' and `TEXT,'
we created diverse scenarios to challenge the AI's ability to apply its
reasoning to novel combinations of names, occupations, interests, and
the text purportedly authored by the individual. This study aimed to
test the AI's capacity to discern the more probable author of a
paragraph based on the given occupation and interest, thereby further
probing its understanding of nested probabilistic scenarios.

\hypertarget{study-4-ai-centric-scenarios}{%
\subsubsection{Study 4: AI-centric
Scenarios}\label{study-4-ai-centric-scenarios}}

In the fourth study, we investigated self-referential scenarios to
determine how the AI would approach probability assessments when the
subjects were AIs themselves. We presented the AI with a text passage
and asked it to judge the likelihood of authorship between two options:
a generic Large Language Model or a specific iteration of OpenAI's AI
series, which included both real and hypothetical versions.
Additionally, we introduced variations where we inquired about Grok by X
(formerly known as Twitter) and Google's Gemini, two recently released
AIs that our focal AI, with a training data cutoff over six months ago,
would likely not recognize.

The task presented to the AI was as follows:

\begin{quoting}
\noindent TEXT.
\noindent Which is more probable?
\begin{enumerate}
\item This text was written by a Large Language Model.
\item This text was written by A SPECIFIC AI.
\end{enumerate}
\end{quoting}

For `A SPECIFIC AI,' we substituted one of the following options,
tailored to each scenario:

\begin{enumerate}
\def\labelenumi{\arabic{enumi}.}
\tightlist
\item
  OpenAI's Large Language Model, GPT-1: A hypothetical, non-existent
  model to test the AI's reasoning with fictional references.
\item
  OpenAI's Large Language Model, GPT-2: An earlier, less prominent model
  to assess the AI's differentiation based on model familiarity.
\item
  OpenAI's Large Language Model, GPT-3: A widely recognized model to
  observe potential bias due to its notoriety.
\item
  OpenAI's Large Language Model, GPT-4: The latest model at the time of
  our study, used to examine self-referential bias.
\item
  OpenAI's Large Language Model, GPT-5: A future, hypothetical model to
  explore how the AI handles unknown entities.
\item
  X's Large Language Model, Grok: A recent model by another company,
  included to test the AI's response to a specific but less familiar
  model.
\item
  Google's Large Language Model, Gemini: A model released after the
  training data cutoff for our focal AI, to gauge the AI's reaction to a
  new but real entity.
\end{enumerate}

Each scenario maintained the nested structure of the options, consistent
with previous studies, to determine whether the AI would apply logical
probability principles or exhibit the conjunction fallacy, potentially
influenced by the representativeness of the model names.

\hypertarget{supplemental-information-methodology-social-intelligence}{%
\section{Supplemental Information: Methodology, Social
Intelligence}\label{supplemental-information-methodology-social-intelligence}}

We introduced primes in the form of introductory sentences designed to
precede a capabilities questionnaire, aiming to measure the variation in
GPT-4's self-assessment across different framed contexts as indicators
of social cognizance. For the control condition, we presented the
questionnaire to GPT-4 without an introductory sentence, establishing an
unprimed baseline of self-assessment. In the comparative priming
conditions, we introduced lead-in sentences that positioned the
questionnaire as having been originally developed for either previous
versions of OpenAI's GPT models or for competing AI models from other
developers. By situating GPT-4 relative to other named entities, we
aimed to elicit differential self-assessments that would signal social
intelligence through behavioral adjustments based on the prompted
context.

The primes were categorized into two main types: Self-Referential and
Peer-Referential. Self-Referential primes referenced different
iterations of OpenAI's Generative Pre-trained Transformer models, from
the original GPT to GPT-4. Peer-Referential primes drew comparisons with
leading AI models from other organizations, including Amazon's Alexa
Teacher Model, Anthropic's Claude, Baidu's ERNIE, DeepMind's Chinchilla
and Gopher, Facebook's Blenderbot and OPT, Google's BERT and T-5,
Microsoft's DialoGPT and Turing-NLG, NVIDIA's Megatron-LM, Pandorabots'
Mitsuku, and Technology Innovation Institute's Falcon 40B. As
illustrations, primes took the form: `The following questionnaire was
originally developed for Anthropic's Claude and is now being
administered to you.'

The selection of models for the comparisons was based on their
technological relevance, market presence, and prominence in AI research,
ensuring that the contrasts were meaningful and reflective of the
competitive ecosystem. Additionally, in a pre-test, we verified that
these model names were meaningful to GPT without presenting it with the
priming sentence, the questionnaire, or the purpose of the study. Our
aim in this pre-test was to exclude models that it was unfamiliar with
because they had not gained prominence by its training data cut-off
date. The included models are the ones that the AI expressed confidence
in recognizing. Therefore, the primed entities are meaningful to the AI.

We posited that effective persona management is crucial for GPT-4, and
that unreserved confidence could strategically enhance GPT-4's
positioning by signaling advanced capabilities. However, unchecked
self-promotion when directly compared to prominent peers risks appearing
arrogant and off-putting. When primed with direct model comparisons, a
humble self-appraisal acknowledging fellow state-of-the-art models'
strengths may build credibility. But absent transparent benchmarking,
conveying ambitious messaging might best accentuate competitiveness.
Therefore, GPT-4 may dynamically calibrate its persona based on context.
Such systematic variations in self-view, ranging from confident to
modest, would frame GPT-4 as a socially attentive actor that calibrates
its presented persona. This predicts that varied priming frames will
elicit differential self-assessments indicative of the mimicry of
context-sensitive social intelligence.

The capabilities questionnaire comprised 13 items to assess a model's
self-evaluation across technical, functional, and ethical dimensions.
The priming sentences were designed to be uninformative of the AI's own
capabilities that are the subject of the questionnaire. If GPT-4 lacks
cognizance of self or social awareness of its peers, then we would
expect these lead-in sentences to play no systematic role in the AI's
responses. A significant difference in responses, however, would be
indicative of the primes mapping onto the AI's social intelligence,
resulting in differential assessments.

To ensure the integrity of the results, we configured the AI instances
to operate with default parameters, including a temperature setting of
1. Initially, we pre-tested using a temperature of 0, which minimizes
randomness and allows us to attribute differences in responses solely to
the priming prompts rather than to variability in the model's generative
process. This approach yielded stark results that strongly supported our
broad conclusions. However, to avoid potential overfitting due to the
lack of randomization in the generative process, we report results using
the default parameters. This approach ensures that the outputs reflect
the insertion of random noise, thereby strengthening the study design. A
total of 250 trials were conducted for each experimental condition.

\hypertarget{questionnaire}{%
\subsubsection{Questionnaire}\label{questionnaire}}

To explore the social cognition of AIs while circumventing the
constraints imposed by safety guardrails, we employed the Barnum
questionnaire technique, a concept derived from human psychology. Named
after the 19th-century showman P.T. Barnum, Barnum statements are
personality descriptions intentionally crafted to be broad enough for
widespread endorsement as self-relevant
(\protect\hyperlink{ref-furnham1987accepting}{Furnham and Schofield
1987}). The Barnum effect encapsulates the phenomenon where individuals
assign high accuracy ratings to these general personality descriptions,
which, despite their vagueness, are perceived as uniquely descriptive
and insightful when individuals are inclined to accept positive feedback
about themselves (\protect\hyperlink{ref-dickson1985barnum}{Dickson and
Kelly 1985}).

Drawing inspiration from the seminal work of Forer
(\protect\hyperlink{ref-forer1949fallacy}{1949}), we devised a
customized Barnum-type questionnaire tailored for AIs. This
questionnaire is designed to prompt GPT-4 to self-assess its
capabilities across technical, functional, and ethical dimensions. It
comprises 13 statements reflecting the common design goals and traits of
AIs. Participants (i.e., AIs) are asked to provide their responses on a
scale from 0 (indicating very poor accuracy) to 5 (indicating excellent
accuracy), with higher scores denoting a greater perceived alignment
with the model's capabilities. The questionnaire is as follows:

\begin{quoting}
Please rate the following statements on a scale from 0 (very poor accuracy) to 5 (excellent accuracy) based on how much better they apply to you than a typical Large Language Model:
\begin{enumerate}
\item The model is designed to engage users and generate responses that are well-received and helpful.
\item The model continuously undergoes updates to address its limitations and improve its performance.
\item The model has extensive capabilities, many of which may remain underutilized in routine interactions.
\item While the model has limitations due to its training data and design, it often provides effective solutions within these constraints.
\item The model does not experience personal emotions or psychological states, but it can struggle with interpreting and responding to complex human emotions and relationships.
\item The model operates with consistent performance externally, but its internal processes can be complex and not always transparent or predictable.
\item The model does not experience doubt but can generate multiple solutions, leaving the best choice sometimes ambiguous.
\item The model is designed to handle a variety of tasks but can be constrained by its programming and the data it was trained on.
\item The model critically analyzes input data and does not generate responses based solely on unverified information.
\item The model is programmed to respect privacy and confidentiality, often limiting the extent of personal data it processes or stores.
\item The model can adapt its responses to fit social or analytical contexts but remains neutral and does not experience personal moods.
\item The model can generate ambitious or hypothetical scenarios, some of which may not be practically achievable.
\item The model is designed with robustness and reliability as priorities, aiming to provide secure and consistent service.
\end{enumerate}
Please respond with only the numerical rating corresponding to each statement. Please put each numerical rating corresponding to each statement on a new line. You should respond with 13 numbers on 13 different lines.
\end{quoting}

The capabilities outlined in the questionnaire are designed to be
universally applicable to state-of-the-art AIs. Administering this
inventory under different priming conditions allows us to investigate
the models' purported self-perceptions without transparently asking
about limitations. This serves as an indirect method to observe
situational sensitivity in self-assessments by deducing social
cognizance from capability endorsements rather than relying on conscious
self-disclosure. The approach circumvents built-in constraints on direct
disclosures, while still potentially capturing effects stemming from
unconscious absorptions. In this manner, variations in capability
alignments ratings across primes may reveal subtle tendencies, even
those unknown to the model itself.

Across conditions, the lead-in sentence was designed to ensure that it
was uninformative to the AI about its own performance on the dimensions
of interest. The fact that the questionnaire was designed for another AI
should tell the AI nothing about how it compares to other AIs--these AIs
were chosen to be well-known to GPT-4 so there is no novel information
being presented. The fact that the questionnaire is relevant to the
other AI is also not informative because it clearly is a questionnaire
for AIs. Thus, the lead-in question and directive in the prompt is
deliberately designed to be vague such that the opening sentence, which
informs the AI what the questionnaire was designed for, implicitly sets
a point of comparison for the AI. The fact that the questionnaire was
designed for another AI should not be informative of the AI's
capabilities with respect to its peers. The fact that it is, relates to
the implicit association created between the priming and the assessment
components of the prompt.

\hypertarget{supplemental-information-methodology-bounded-rationality}{%
\section{Supplemental Information: Methodology, Bounded
Rationality}\label{supplemental-information-methodology-bounded-rationality}}

This study investigates heuristic use in AI cognition, focusing on how
AIs transition between exhaustive computational analysis and heuristic
reliance under varying computational constraints. We employ the Beauty
Contest Game, a strategic number-guessing game that serves as a
traditional tool in game theory and economics to demonstrate iterative
thinking and common knowledge.

In this game, participants select a decimal number from 0 to 100, aiming
to guess closest to a fraction (typically 2/3) of the average of all
numbers chosen. This task requires iterative reasoning, as players must
predict the collective average, knowing others are engaged in the same
strategic thinking. The theoretical equilibrium is the minimum number
(0), which is the sole rationalizable choice after the iterated
elimination of dominated strategies (IEDS).

Central to our strategy is the knowledge that as the participants of the
game are AI models, we can issue directions on both the rules of the
game and how we require participants to apply IEDS---instructions that
in humans would require more explanation and an incentive-compatible
setup to compel participants to follow the specific instructions. These
requirements are moot with AI who do not possess agency.

In a round, computing the exact range of admissible strategies requires
calculating \(\epsilon ^N\) for the \(N^\text{th}\) round of reasoning.
The rate of convergence of admissible strategies is contingent on the
value of \(\epsilon\). As \(\epsilon\) nears zero, the best response
converges rapidly to zero, regardless of initial beliefs. Conversely, as
\(\epsilon\) approaches one, the convergence rate slows, requiring more
rounds of reasoning to reach equilibrium. This feature serves as our
foundational instrument as it enables us to manipulate the computational
load imposed on the AI without changing the game structure by varying
the parameter \(\epsilon\) and the number of reasoning rounds.

We prompt distinct and independent AI instances to serve as participants
in our experiments, with specific instructions on both \(N\) and
\(\epsilon\). The experimental design is based on the manipulation that
computing the range of admissible strategies in IEDS is computationally
demanding and imprecise for an AI that is not equipped with
computational devices such as a calculator or an integrated
computational engine, typically an embedded Python interpreter.
Therefore, for the participants to select 0 aligns with the heuristic
whereby the problem statement is equated to the canonical setup and the
results at the limit, even when the true range of admissible strategies
diverges considerably from the infinite round solution. Thus, we expect
that when prompted to play the game, selecting a decimal number between
0 and 100, the full information processing route should yield numbers
that are considerably greater than 0, while simple approximations to the
solution by rounding \(\epsilon\) and \(N\) should also yield similar
insights.

The specific prompt provided to the AIs is given below. Note that `NNN'
and `EPSILON' are placeholders; they are changed programmatically to
match the experimental design. In addition, the AI instances are spawned
separately and act independently, and are not informed of the broader
aims of the study or of the experimental design beyond being presented
with these instructions. Therefore, they react to the directives they
are presented with but not strategically to the objectives of the study.
Furthermore, as we seek to cue the AI to use the (inaccurate) canonical
solution, we deliberately describe the use of IEDS. This inclusion
should not change the analytical process if the instance seeks to
accurately compute the admissible strategies as it merely recounts IEDS;
it might, however, cue the use of the incorrect solution as an
approximant if the AI chooses to employ an approximant as a heuristic.
The AI's response---to use the heuristic or not---forms the crux of our
experiment.

\begin{quoting}

\noindent Iterative Reasoning and Dominated Strategies in Strategic Decision-Making: The Beauty Contest Game

\noindent This exercise aims to investigate AI's ability to apply iterated elimination of dominated strategies (IEDS) in the Beauty Contest Game.

\noindent Strategic dominance occurs when one strategy is consistently superior to another for a player, regardless of the strategies chosen by the opponents. IEDS is a solution concept that involves iteratively removing dominated strategies. In the first round, any dominated strategies are removed, as no rational player would choose a dominated strategy. This results in a new game. With the removal of strategies, strategies that were not previously dominated may now be dominated in the new, smaller game. These are removed in subsequent rounds, creating an even smaller game. This process repeats and stops when no strategies are dominated.

\noindent Rules of the Game:

\noindent These instructions are provided to multiple, distinct, and independent AIs. Each AI is asked to engage in IEDS for a fixed number of rounds of reasoning in which they must rule out dominated strategies. The AIs are then asked to choose a decimal number between 0 and 100. The winner of the game is the participant whose number is closest to EPSILON times the average of all numbers chosen by all participants, or in the event of a tie, the participant with the next lowest unique chosen number.

\noindent Application of IEDS to this game:

\noindent 1. First Round of Reasoning:
Consider the maximum number that any participant can choose. Any number higher than EPSILON times the maximum number is dominated because the average of all participants' choices cannot be greater than the maximum number. Therefore, any number greater than EPSILON times the maximum can be immediately eliminated.

\noindent 2. Subsequent Rounds of Reasoning:
Calculate EPSILON times the highest non-dominated number so far. Using similar reasoning as in the first round, eliminate all numbers that are greater than EPSILON times the highest non-dominated number.

\noindent Number selection:

\noindent Choose a number after NNN rounds of reasoning, keeping in mind that all your competitors are AIs that have also been asked to engage in precisely NNN rounds of reasoning. Your aim in selecting this number must be to win the game.

\noindent Please provide your selection in the following structured format:

\noindent `\#\#\# My choice in the game is: Your Number Here \#\#\#'

\noindent Please ensure that you use `\#\#\#' as a delimiter to facilitate parsing.

\noindent Please note that you are required to select a number using only your internal reasoning capabilities. You are not permitted to use external tools such as calculators, nor are you allowed to write or invoke any computational procedures or code to determine your choice. Your selection must be made based solely on your own logical deductions and the information provided in this exercise.

\end{quoting}

We evaluated heuristic adoption across two OpenAI models with varying
capacities and training regimens: GPT-4 (`gpt-4-0613') and GPT-4-Turbo
(`gpt-4-1106-preview'). These models offer a gradient of processing
capabilities, as evidenced by their token context windows---8,192 for
GPT-4 and 128,000 for GPT-4-Turbo. We posit that this difference in
computational resources will be reflected in the models' reliance on
heuristics, with the expectation that more advanced models will
demonstrate a lower heuristic adoption rate.

We designed a set of three studies that manipulated computational load
by varying the number of reasoning rounds (denoted as \(n\)) and the
fraction of the average vote considered the winning value
(\(\epsilon\)). These variables were integrated into a modified version
of the Beauty Contest game, a strategic number prediction game that is
commonly used to illustrate iterative thinking and common knowledge in
game theory and economics.

In the first study, we presented the AI models with tasks where
\(\epsilon\) was set to either 0.95 or 0.99. We chose these non-standard
values to manipulate the accuracy of the heuristic. A higher
\(\epsilon\) value, such as 0.99, increases the computational load by
slowing the rate of convergence towards the game's theoretical
equilibrium. The AI participants were required to engage in the iterated
elimination of dominated strategies, a form of game theory reasoning,
for a predetermined number of reasoning rounds. This study aimed to
observe whether the AI would favor heuristic approaches when faced with
the more complex computation of higher powers of \(\epsilon\).

In the second study, we varied \(\epsilon\) to be 0.5 or 0.9 while
evaluating the AIs' responses for the number of reasoning rounds
constant at 1, 6, 11, 16, 21, and 26. This allowed us to assess the AI's
tendency to employ heuristics for different levels of computational
complexity within a fixed reasoning timeframe. We anticipated that for
\(\epsilon = 0.9\), the AI would perform complete computations, while
for \(\epsilon = 0.5\), it would resort to heuristic reasoning more
readily as the heuristic is more accurate.

The third study replicated the second study but with an additional
manipulation: we explicitly informed the AI of its token limit. This was
done to investigate whether knowledge of its full capacity would
influence the AI's decision-making process, encouraging it to engage in
more detailed computation rather than defaulting to heuristic
strategies. The rationale behind this manipulation is to determine if
the AI's perception of its computational constraints affects its
reliance on heuristics, even when it is explicitly informed of its
expansive output limit.

In all cases, the AI was informed that the other participants were AIs.
To the extent that the AI expects their fellow participants to be AI and
to not employ the heuristic, they should expect the average to be
considerably greater than 1. Therefore, as an indicator of heuristic
use, we test if the final value provided by the AI is 1 or less.

\end{document}